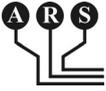

# New Intelligent Transmission Concept for Hybrid Mobile Robot Speed Control

**Nazim Mir-Nasiri & Sulaiman Hussaini**
Department of Mechatronics, Faculty of Engineering
International Islamic University Malaysia, Kuala Lumpur, Malaysia
nazim@iiu.edu.my

*Abstract:* This paper presents a new concept of a mobile robot speed control by using two degree of freedom gear transmission. The developed intelligent speed controller utilizes a gear box which comprises of epicyclic gear train with two inputs, one coupled with the engine shaft and another with the shaft of a variable speed dc motor. The net output speed is a combination of the two input speeds and is governed by the transmission ratio of the planetary gear train. This new approach eliminates the use of a torque converter which is otherwise an indispensable part of all available automatic transmissions, thereby reducing the power loss that occurs in the box during the fluid coupling. By gradually varying the speed of the dc motor a stepless transmission has been achieved. The other advantages of the developed controller are pulling over and reversing the vehicle, implemented by intelligent mixing of the dc motor and engine speeds. This approach eliminates traditional braking system in entire vehicle design. The use of two power sources, IC engine and battery driven DC motor, utilizes the modern idea of hybrid vehicles. The new mobile robot speed controller is capable of driving the vehicle even in extreme case of IC engine failure, for example, due to gas depletion..
*Keywords:* hybrid drive, intelligent gear transmission, fuzzy logic.

## 1. Introduction

Variety of controllers have been developed to improve driving as well as operating conditions of vehicles, such as to provide smooth throttle movement, zero steady-state speed error, good speed tracking over rod slopes, and robustness to system variation and operating conditions. The complexity of the speed control algorithms has increased through the years to meet the growing demands on more stringent automotive performance criteria. The earliest systems simply held the throttle in a fixed position [1]. In the late 1950's speed control systems with feedback immerged [2]. These used proportional feedback of the speed error to control the position of the throttle. The next enhancement was proportional control with integral preset or bias input [3]. This minimized steady-state error as well as speed drop when the system was initialized. Only with the recent availability of inexpensive microprocessors more sophisticated control strategies have been implemented. PID controllers, optimal LQ regulators, Kalman filters, fuzzy logic have all been tried [4]-[10]. Some researches [11] developed adaptive speed controllers to meet simultaneously all the objectives mentioned above which is impossible by conventional fixed gain controllers. To reduce the complexity, the controller design is based on sensitivity analysis and slow adaptation using gradient methods. The adaptive algorithm in this design driven by the vehicle response to road load torque disturbances, tunes a PI controller to continuously minimize a single performance based cost functional for each different vehicle over varying road terrain. All designed speed controllers so far were aimed to improve driving conditions of a car, for example by smooth throttle control, irregardless of its gear transmission. However, the transmissions play a vital role for the normal working of a vehicle. Firstly, an IC engine provides maximum torque in a limited band of speeds. Secondly, its maximum speed is limited, usually by the valve train. To overcome these factors the transmission allows the gear ratio between the engine and the drive wheels to change as the car speeds up and slows down. There are two main types of transmission in cars: manual transmission and automatic transmission. The key difference between a manual and an automatic transmission is that the manual transmission locks and unlocks different sets of gears to the output shaft to achieve the various gear ratios, while in an automatictransmission the same set of gears produces all of the different but discrete gear ratios. The planetary gear set is the device that makes this possible in an automatic transmission. However, there are a number of shortcomings of the above mentioned types of transmissions. Manual transmissions can be very tiring to drive in heavy stop-and-go traffic. It has the problem of worn or slipping clutches. In cold weather manual transmissions can become sluggish and hard to shift. Also in the beginning can be difficult to learn. The automatic transmission has its demerits as well. They tend to reduce the overall pick up of the vehicle to a considerable extent. It is very heavy or bulky. The fuel consumption is far more than a manual one. During steep descent the slip in the torque converter fails to transmit engine braking effectively. The driver is not in full control over the drive train and the engine's power output to the wheels. Also it is more costly than the manual transmission.

In the new approach we use the same epicyclic gear train but in different function, to monitor and control the car's driving and operating conditions. The gear train has two inputs that are coupled with the shafts of the IC engine and dc motor respectively. Hence, both IC engine and battery driven DC motor are simultaneously involved in driving the vehicle. By intelligently changing the speeds of two inputs all possible car driving conditions, such as idling, acceleration, deceleration, stop, reverse can be achieved. The gear ratio of the epicyclic gear train is selected to reduce the load on the DC motor shaft compared to that of the IC engine. It approach allows to prolong lifespan of the battery in such hybrid car. The implementation of the new design enables to overcome many problems encountered in traditional cars. The overall performance of the vehicle can be enhanced significantly.

## 2. Conceptual Design and Model ling of the System

*2.1 Epicyclic Gear Train Selection and Analysis*
Most automotive automatic transmissions provide four forward speed ratios and one reverse speed ratio [12]. The schematic diagram of elementary planetary, or epicyclic, gear train is shown in Fig. 1. It has been selected for the design of the intelligent speed controller.

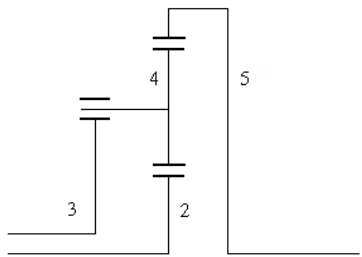

Fig. 1. Schematic diagram of two-DOF epicyclic gear train

The elementary train consists of two gears, the sun (1) and planet (2) gears, and a third member, hereafter referred to as the planet carrier or arm (3).The following is the elementary derivation of the formula that constitutes the speed relations between two inputs (sun gears 2 and 5) and one output (arm):

$$\frac{\omega_5 - \omega_{arm}}{\omega_2 - \omega_{arm}} = -\frac{N_2}{N_5}, \quad (1)$$

where $N_i$ denotes the number of teeth of the $i^{th}$ gear ($i = 2, 5$), $\omega_i$ denotes the speed of the $i^{th}$ gear ($i = 2, 5$), $\omega_{arm}$ denotes the speed of the arm link. Modifying equation (1) leads to:

$$(\omega_5 - \omega_{arm})N_5 = -(\omega_2 - \omega_{arm})N_2$$
$$\omega_5 N_5 - \omega_{arm} N_5 = -\omega_2 N_2 + \omega_{arm} N_2$$
$$w_{arm}(N_5 + N_2) = \omega_5 N_5 + \omega_2 N_2$$
$$\omega_{arm} = \omega_5 \left(\frac{N_5}{N_5 + N_2}\right) + \omega_2 \left(\frac{N_2}{N_5 + N_2}\right) \quad (2)$$

Since $\left(\frac{N_5}{N_5 + N_2}\right) > \left(\frac{N_2}{N_5 + N_2}\right)$ in (2), the gear 2 is

selected for coupling with the engine shaft, the gear 5 is selected for coupling with the DC motor shaft, and the arm link of the gear transmission is selected for coupling with the vehicle wheels. Obviously $\omega_2$, $\omega_5$ and $\omega_{arm}$ are engine, DC motor, and wheel angular speeds, respectively. Since the speed $\omega_{arm}$ is a linear combination of speeds $\omega_2$, $\omega_5$ the contribution of the DC motor speed ($\omega_5$) to the total speed of the wheels is $N_5 / N_2$ less than that of the engine speed ($\omega_2$). This selected arrangement of gears will enable somewhat to save power of the battery driving the DC motor during the ride of the mobile robot.

*2.2 Intelligent Controller Design*
The intelligent speed control system should be designed to provide smooth ride and robustness of the system to varying operating conditions. In this work a controller has been designed to vary the speed of the vehicle for different driving conditions. The block diagram which identifies all necessary functional relations between the controller and other subsystems of the mobile robot is shown in Fig. 2.

The controller is used as an interface between the vehicle driver and two power sources, IC engine and DC motor. Driver controls the vehicle by pressing either accelerator or brake pedals.

The controller responds to the driver commands and selects an optimal driving condition for the car vehicle mixing intelligently the energies (speeds) from the two power sources by means of epicyclic gear train. The feedback loop is used additionally to monitor the actual speed of the DC motor. All basic driving conditions for the car have been identified as follows:

i) *Vehicle idling* -engine runs, car is still stationary, both pedals are untouched
ii) *Vehicle accelerates* – accelerator pedal is being pressed, brake is not pressed
iii) *Vehicle runs with constant speed* – no change in position of the accelerator pedal.
iv) *Vehicle decelerates* – brake pedal is being pressed, accelerator is not pressed
v) *Vehicle stops* – brake pedal has been pressed for long time, no accelerator pedal
vi) *Vehicle reverses* – reverse switch is on, accelerator is being pressed, no brake.

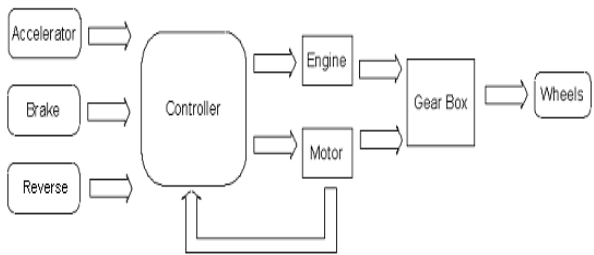

Fig. 2. Functional interrelations between subsystems

The new controller is implemented by using fuzzy logic and simulated in Simulink (MATLAB 7). The apparent success of fuzzy logic controller (FLC) can be attributed to its ability to incorporate expert information and generate control surfaces whose shapes can be individually manipulated for different regions of the state space with virtually no effects on neighboring regions. FLC is ideal for the speed control problems, since there is no complete mathematical model of the engine and other components of the car. However, some human driving experience and visual feedback can be used in the design of control system as well. Human operators control the speed of the car by pressing the accelerator pedal, which opens and closes the throttle valve of the engine. In addition to this the brake pedal and stop button are also introduced to reduce the speed and stop the car if necessary. From these human actions, fuzzy rules were formulated using the amount and the rate at which the accelerator and the brake pedals are pressed. The membership functions to represent the inputs and the outputs of FLC are symmetric triangles with equal distribution over the entire range or the universe of discourse. For example, the input values to the controller from the accelerator pedal are divided into four membership functions indicating the rate of change of pedal position. They are zero, slow, medium and fast rate values. Similar functions are developed for the brake pedal. In the design of controller all the inputs and outputs are presented in the form of voltage signals. Therefore, the output membership functions are named as 'Idle voltage', 'zero voltage', 'low voltage', 'medium voltage' and 'high voltage'. There are total of 24 rules formulated for the controller design using Mamdani implications.

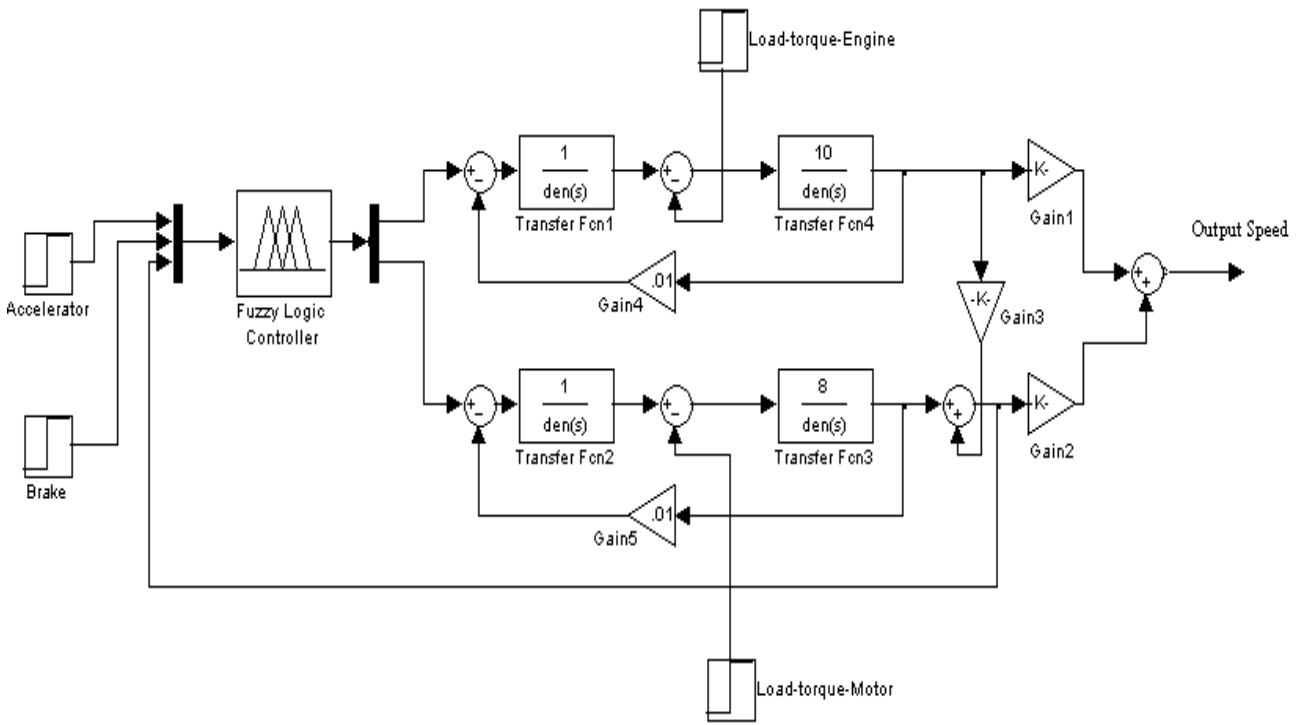

Fig. 3. Operational block diagram of the speed control system

The block diagram of the proposed system for the simulation is shown in the Fig. 3. The vehicle engine in simulation work has been replaced by the second armature controlled DC motor. Three gains of the block diagram (Gain1, Gain2 and Gain3) are reflections of the characteristic equation of the selected epicyclic gear train (2). The FLC block decides on voltage inputs for two DC motors coupled with two inputs of the epicyclic gear train. FLC makes decision based on the signals received from the potentiometers of the accelerator, brake pedals and the feedback signal from one of the DC motor shaft's optical encoder.

## 3. Simulation Results

Table 1 shows parameters of the system components that have been selected used for simulation studies. The details of closed-loop armature speed control of each DC motor are shown in Fig. 4. The simulation was conducted for basic driving conditions of the vehicle.

*3.1 Idle Run of the Engine*
The results of simulation are shown in Fig. 5. This figure shows three zones of speed control:

A) The system is off
B) The system is on, no pedals are pressed
C) The engine idles with constant speed

| DC motors | $J = 0.01$ N.m.s$^2$/rad<br>$b = 0.1$<br>$k_m = 10$ N.m /A   (Engine- motor constant)<br>$k_m = 8$ N.m /A     (DC motor constant )<br>$R = 1$ ohm.<br>$L = 0.5$ F<br>Load = 0.001N.m |
|---|---|
| Epicyclic gear train | $N_5$ =71 (number of teeth of ring gear)<br>$N_4$ =16 (number of teeth of planet gear)<br>$N_2$ =39 (number of teeth of sun gear)<br>The characteristics equation with gains:<br>(Gain1 = 0.35, Gain2 = 0.64, Gain3= -0.55)<br>$\omega_{arm}$ =0.6454 $\omega_5$ +0.3545 $\omega_2$ |

Table 1. System Parameters

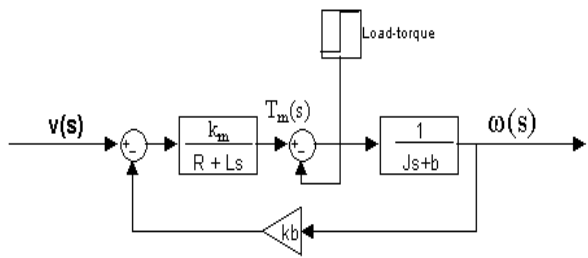

Fig. 4. Closed-loop armature speed control of DC motor

In the zone B the engine is energized and picking up the speed. The DC motor is still off. Since the output of the gear train (coupled with the wheels) is loaded with the weight of the car it acts as an input with zero speed and the DC motor shaft acts then as an output gradually increasing speed in negative direction. In the zone C, the speed of the wheel is zero and the speeds of the engine and DC motor shafts are stabilized and strictly follow the equation (2). This equation is implemented in the block diagram of the system (Fig. 3) and controlled by the set of gains (Gain1, Gain2 and Gain3).

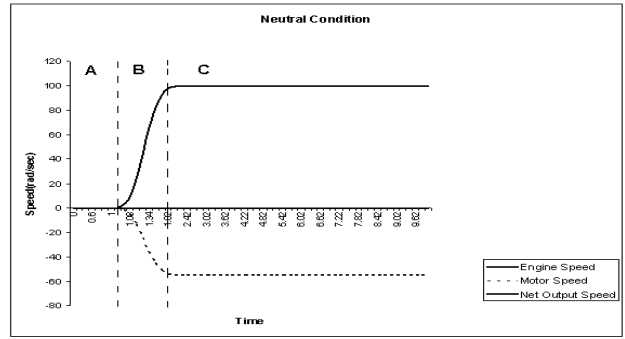

Fig. 5. Idle run of the engine

*3.2 Acceleration, Deceleration, and Stop of the Vehicle*
The results of simulation are shown in Fig. 6 and Fig. 7. Zone A in Figure 6 shows the engine idling condition with no moving wheels. Zone B shows the response of the system in terms of three speeds ($\omega_2$, $\omega_5$ and $\omega_{arm}$) to the ramp input applied to the accelerator pedal.

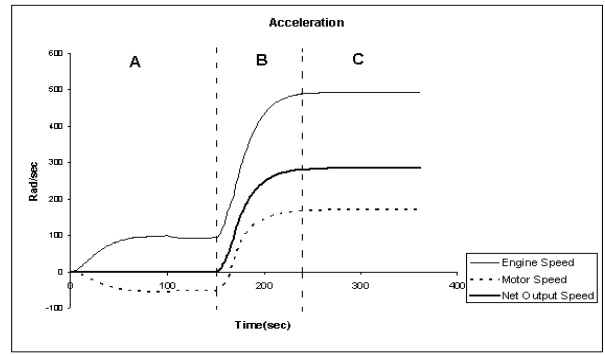

Fig. 6. Acceleration of the car

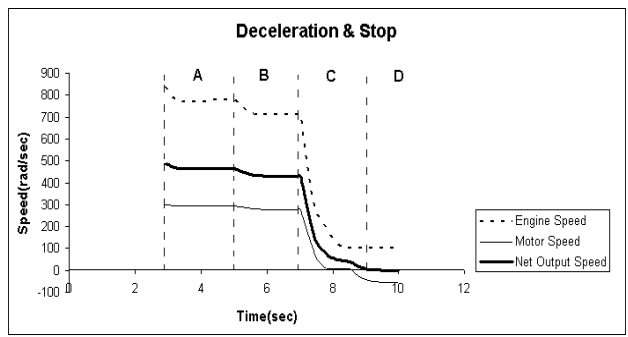

Fig. 7. Deceleration of the car

Zone A in Figure 7 shows the response of the system to the condition when the brake is applied (ramp signal) for sometime (pressed and released). In zone B the brake again pressed and released. Finally, in zone C the brake is pressed until the vehicle comes to a halt. Zone D shows the vehicle in the neutral condition (engine idles). The advantage of the system is that there is no physical brake in the system and the vehicle stop condition is fully implemented by controlling the speed of the DC motor only. The DC motor forces the engine *to return to idle condition* (2) at the end of deceleration in order to provide zero speed for the wheels.

*3.3 Reverse of the Vehicle*

The results of simulation are shown in Fig. 8. The reverse is initiated by pressing an additional reverse switch. At this condition the controller increases the negative speed of the DC motor in order to exceed initially balanced positive speed of idling engine shaft.

As the result the wheels will gradually move in negative direction (Fig. 8). The more is the speed of the DC motor the more is the speed of reversing vehicle.

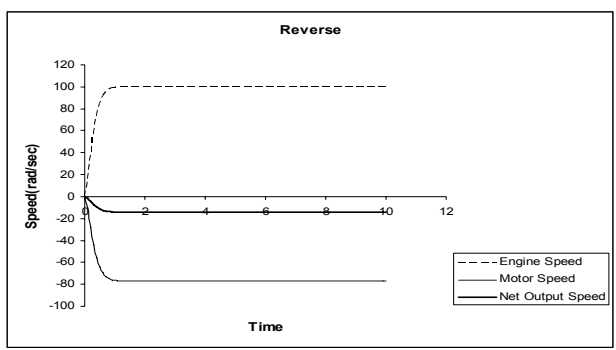

Fig. 8. Reverse of the car

## 4. Conclusion

The paper presents a new control approach in driving a vehicle or mobile robot The controller is built on the speed mixing capability of a two degree of freedom epicyclic gear train. Signals from the accelerator, brake pedals and reverse button are intelligently treated in FLC to generate input signals for two driving actuators – car engine and additional DC motor. They, in turn, jointly control the speed of vehicle wheels according to the characteristic equation of the selected epicyclic gear train. This design successfully utilizes a new idea of hybrid vehicle recently immerged in automotive industry. The system does not require a physical braking subsystem which will reduce the overall cost of a car. This paper illustrates the simulation results of basic driving conditions of a car, such as engine idling, car acceleration, deceleration, stop, and reverse. In case of the autonomous engine driven mobile robot the signals from the pedals are to be replaced by the signals from the central processing unit of the robot for controlling its speed.